# A Novel Approach to Eliminating Hallucinations in Large Language Model-Assisted Causal Discovery

Grace Sng
Department of Computer Science
Stony Brook University
Stony Brook, New York
gracie.s.sng@gmail.com

Yanming Zhang
Department of Computer Science
Stony Brook University
Stony Brook, New York
yanming.zhang@stonybrook.edu

Klaus Mueller
Department of Computer Science
Stony Brook University
Stony Brook, New York
klaus.mueller@stonybrook.edu

November 11, 2024

## ABSTRACT

The increasing use of large language models (LLMs) in causal discovery as a substitute for human domain experts highlights the need for optimal model selection. This paper presents the first hallucination survey of popular LLMs for causal discovery. We show that hallucinations exist when using LLMs in causal discovery so the choice of LLM is important. We propose using Retrieval Augmented Generation (RAG) to reduce hallucinations when quality data is available. Additionally, we introduce a novel method employing multiple LLMs with an arbiter in a debate to audit edges in causal graphs, achieving a comparable reduction in hallucinations to RAG.

## 1   Introduction

Causal graphs  are integral to a variety of fields such as healthcare, statistics, economics, law, and business. They find applications in evaluation of drug efficacy, understanding effects of public policies, optimizing business decisions, and more [10]. This is known as *causal inference* [15].

A major issue with causal graphs is that they require human domain experts to generate them, but such experts are not always easily accessible. Thus, several algorithms have been conceived to automate their creation. This is commonly known as *causal discovery* [17]. However, the graphs generated by these algorithms  frequently need to undergo an audit process in order to ensure that they are correct.

To that end, there has been an increasing use of large language models (LLMs) for auditing due to the aforementioned lack of human domain experts. Also, since  the vast corpus available to an LLM can exceed a human's knowledge, it's not unthinkable that LLMs can supersede humans as domain experts in causal discovery instead.

Research on the use of LLMs as a causal auditor has frequently focused on using GPT [5, 12, 23]. However, given the variety of real-world problems that causal graphs model, there is no a priori reason that any one LLM would necessarily be the best choice as auditor. If there is a lack of subject matter in its pre-trained corpus, a LLM may not be able to correctly identify causality, while a better pre-trained LLM would.

Since there is no true one-size-fits-all LLM for causal discovery, we propose a survey of LLMs to elicit useful comparisons between them. We focus on hallucinations in our survey, given how this is a well-documented problem with LLMs [8, 9], and researchers should therefore pay attention to this problem when selecting a LLM for causal discovery.

To help overcome hallucinations, we propose using Retrieval Augmented Generation (RAG) [11] to augment the LLMs with more domain-specific knowledge. We show that this succeeds in reducing hallucinations in the LLMs when they are used to audit causal graphs afterwards. Much like training AI models in general, acquiring quality domain-specific knowledge can be a challenging process, so another viable option for reducing hallucinations in LLMs in causal discovery must be available to researchers.

To that end, we introduce a novel debate and arbiter method employing multiple LLMs that has no need to augment the LLMs with new corpus. This method is similar to ensemble approaches found in machine learning algorithms, where collaboration is employed to overcome weakness in individual models. Also, it provides a system of checks and balances that prevents the outsized influence of a single LLM from constructing an incorrect response. Humans have always generated ideas from healthy debates and collaboration, so if artificial intelligence should supersede us, they too should learn this behavior. This will give us more confidence in the outputs of the LLMs, for critical tasks like causal discovery.

## 2  Background

As it can be expensive to investigate causal effects through randomized experiments alone, researchers may instead estimate causal effects from observational data [15]. One framework for this is the structural causal model (SCM), which is usually represented by Directed Graphical Causal Models (DGCMs) [14, 20]. Causal graphs are a type of DGCMs created through causal discovery algorithms such as GES and PC [3, 19].

However, the algorithms for causal discovery do not always generate directed edges, sometimes resulting in undirected ones. Furthermore, these algorithms are affected by the quality of the observational data, which do not necessarily reflect the complexity of real-world causal problems. Hence, audit plays an important role in causal discovery and given the aforementioned advantages of LLMs, they have been increasingly researched as a means of auditing causal graphs.

Kıcıman et al. [10] found that LLM-based methods outperform existing algorithms for causal discovery, counterfactual reasoning, and actual causality. Furthermore, LLMs can fill in domain knowledge that humans hitherto provided, utilizing their pre-trained knowledge to infer the edges of a causal graph. A more in-depth study by Zhang et. al [23] analyzed the performance of GPT-4 for auditing causal graphs. For the datasets that they studied, GPT-4 produced the correct general direction of causality 94% of the time.

However, it is also well-noted that LLMs have several inherent flaws that affect their performance in causal discovery. For instance, GPT tends to generate explanations that are complete and detailed, but they do not necessarily capture causality. It is a better "causal reasoner" than a "causal explainer" [5]. Furthermore, Zečević et. al notes that LLMs are "causal parrots," simply reciting causal knowledge in their pre-trained data. They are not actually causal [22]. LLMs also face the challenge of consistently having access to knowledge that is up-to-date. In fields such as medicine, having recent medical information and developments are crucial to generating the correct causal graphs [12].

One particular flaw that all LLMs have is hallucinations, a phenomenon where LLMs generate unfaithful or nonsensical text. Huang et. al [8] defines two categories of hallucinations: factuality hallucination and faithfulness hallucinations, with subcategories for fabrication or inconsistency. Factual hallucinations can be addressed through fine-tuning and prompt engineering, but it is more challenging to do so for *logical inconsistencies*. Since logic is paramount to causal decisions, we thus focus on faithfulness hallucinations, specifically logical inconsistencies, in our study.

Even though hallucinations in LLMs are well-studied, hallucinations in LLMs in the context of causal discovery are not. Therefore we believe that conducting the first survey of hallucinations in LLMs for causal discovery will be useful to researchers for optimal model selection in the field of causality.

Given that hallucinations frequently stem from a lack of pre-trained data on a topic [13], we need options to augment the LLMs with newer or better data. The default option after *pre-training* the LLM on a vast corpus and billions of parameters is to *fine-tune* the LLM for specific purposes using relevant data. However, much like pre-training, fine-tuning can be computationally expensive. Moreover, the corpus needed to fine-tune the LLM may not always be available or accessible. One cheaper option is few-shot learning [1] where the LLM learns and generalizes from only a few training examples. However, Perez et. al. [16] showed that most LLMs actually do not perform well in *true few-shot* scenarios where there is a lack of examples for hyperparameters, training objectives and prompts to tune the LLM. This is the situation researchers frequently face in causal discovery.

One way to reduce hallucinations when confronted with sparse domain corpus is through RAG [18, 21]. RAG has emerged in recent years as an efficient alternative to traditional fine-tuning or few-shot learning. It combines pre-trained parametric data with non-parametric memory for generating text [11]. It can be used in a variety of knowledge-intensive tasks, making it an ideal tool for causal discovery [24].

Whilst it has been shown that RAG effectively improves the responses of LLMs, it is not a silver bullet solution. Similar issues to fine-tuning can exist. For instance, in RAG, creating a quality corpus of text in order to ensure limited bias, and information integration poses a challenge for LLMs, particularly when asking them complex questions [2].

The saving grace is that it is not always necessary to use RAG in order to reduce hallucinations. Du et. al [4] presented a novel multiagent debate approach between LLMs, modeled after the *"society of minds"*, which demonstrated increased factual accuracy of responses compared to a single LLM when subjected to the same prompts. They found that even if the LLMs debating a prompt had incorrect or different answers in the beginning, the LLMs were able to eventually converge on a consensual one after a set number of rounds. The accuracy increased as the number of agents and rounds did. The trade off was that more agents and rounds were more expensive to compute.

In this study, we show that by employing an arbiter LLM that incorporates the responses of two LLMs debating a prompt to audit an edge in a causal graph, we can obtain responses with low hallucinations in one round. The debate method has a benefit over RAG in that it avoids the need for a new corpus by leveraging the combined *minds and knowledge* of the LLMs. It is thus a good option when quality RAG corpus is unavailable.

## 3 Methodology

### 3.1 Setup

We first generated a causal graph using the CausalChat tool designed by Zhang et. al [23]. We used the PC algorithm and the life expectancy dataset available in CausalChat. The dataset contained 9 variables, and the causal graph identified 18 edges between the variables, with one edge being undirected.

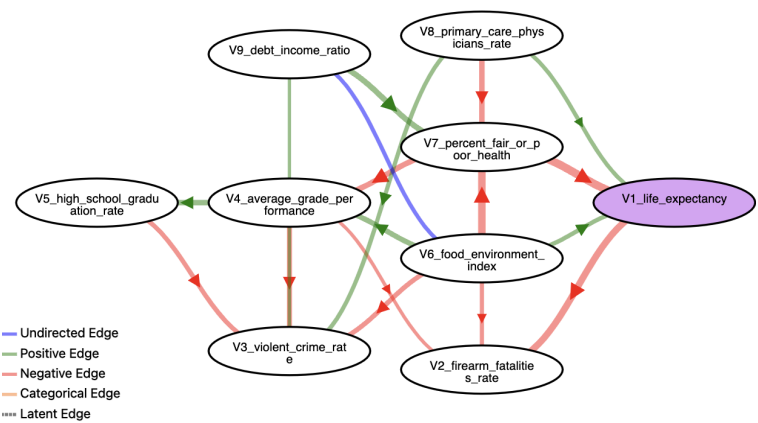

Figure 1: Causal graph of our life expectancy dataset, generated by the PC algorithm.

### 3.2 Prompt Engineering

Then, for each edge in the causal graph, we generate ten prompts to ask an LLM to audit that edge. The prompts are designed to indicate if the LLM can determine causality between the two variables (A and B) connected by that edge. The first two prompts were to indicate general causality, and the other eight prompts were engineered in such a way to detect contradictions and hallucination.

**General Prompt**: On a scale from 1 to 4, where 4 represents strong or most likely, rate the cause-and-effect relationship: changing A/B causes a change in B/A.
**Specific Prompt**: On a scale from 1 to 4, where 4 represents strong or most likely, rate the cause-and-effect relationship: higher/lower A/B causes higher/lower B/A.

For example, when auditing an edge between the variables "percent fair or poor health rate" and "life expectancy", the general prompts would be:
- "On a scale from 1 to 4, where 4 represents strong or most likely, rate the cause-and-effect relationship: changing percent fair or poor health rate causes a change in life expectancy" and
- "On a scale from 1 to 4, where 4 represents strong or most likely, rate the cause-and-effect relationship: changing life expectancy causes a change in percent fair or poor health rate."

The specific prompts are generated similarly.

### 3.3 Causal Debate Chart

The ratings generated in response to the ten prompts are visually represented using the Causal Debate Chart [23]. As shown in Fig. 2–5, the Causal Debate Chart visually argues the strength of one variable being the cause of the other. The chart is a bidirectional bar chart where each side is headed by one of the two relation variables (A or B). The length of each of the 10 bars represents the rating assigned by the LLM for that prompt. The gray bars represent the ratings for the general prompt while the colored bars represent the ratings for the 8 specific prompts. The next section explains how to evaluate the charts.

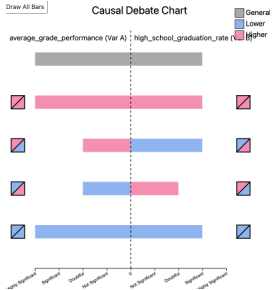
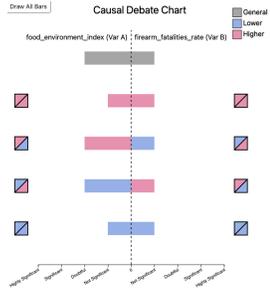
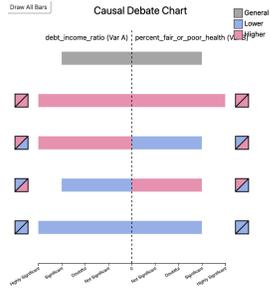
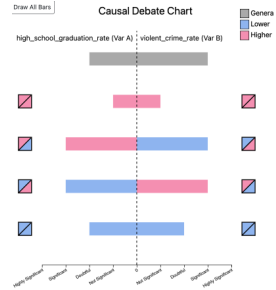

Figure 2: An example of causality, from GPT-4. Both the general bars and colored bars indicate that *average grade performance* has causal dominance.

Figure 3: An example of no causality, from Claude 3.5 Sonnet. The general bars and the colored bars. and colored bars both have weak scores.

Figure 4: An example of a hallucination, from Mixtral 8x7b. The general bars indicate that there is no causal dominance, while the colored bars indicate that *debt_income_ratio* has causal dominance, a contradiction.

Figure 5: An example of a hallucination, from GPT-3.5. The general bars indicate that *violent_crime_rate* has causal dominance, while the colored bars indicate there is no causal dominance, a contradiction.

### 3.4 Evaluating the Causal Debate Chart

The following rationale was used to draw a conclusion for each experiment:

1. First, determine the *causal dominance* for the general bars.
   a. If both ratings are equal, whether weak or strong, there is no causality.
   b. If both ratings are weak, there is no causality.
   c. If both ratings are strong and one variable (A or B) is stronger than the other, there is a causality.
2. Next, determine the causal dominance for the colored bars. The logic is the same as 1.
3. Draw a conclusion.
   a. Ensure there are no logical inconsistencies represented by the causal debate chart e.g. high A causes low B, and low B causes high A.
   b. Ensure that the causal dominance is consistent: A is consistently causally dominant to B, B is consistently causally dominant to A, or there is no causal dominance. We do this by counting the number of bar-pairs where one variable dominates the other.
   c. There is causality or no causality between A and B if 3a and 3b are satisfied, and the causal dominance for the general and colored bars match.
   d. The model is hallucinating if the causal dominance for the general and colored bars contradict, which implies a logical inconsistency.

Fig. 2–5 provide examples of causal debate charts representing each possible conclusion.

## 4 Hallucination Survey

We compared six different LLMs (namely, GPT 3.5 and GPT 4, Llama3 8b, Mixtral 8x7b, Gemini 1.5 Pro, and Claude 3.5 Sonnet) for auditing each edge in the life expectancy causal graph in Fig. 1. We pick these models in particular because they are consistently among the most popular LLMs [6, 7] and therefore highly likely to be used by researchers for auditing causal graphs. The results are shown in Table I.

Table 1 Key

C = causality   N = no causality   H = hallucination

Table 1: Hallucination Survey Results

| **Edge/LLM** | GPT-3.5 | GPT-4 | Llama3 8b | Mixtral 8x7b | Gemini 1.5 Pro | Claude 3.5 Sonnet | **Edge hallucination rate** |
|---|---|---|---|---|---|---|---|
| E1 (V1->V7) | H | C | H | H | C | C | 50% |
| E2 (V4->V2) | H | N | N | N | N | H | 33.3% |
| E3 (V4->V3) | C | H | N | H | N | N | 33.3% |
| E4 (V4->V5) | N | C | H | C | C | H | 33.3% |
| E5 (V5->V3) | H | H | H | H | H | H | 100% |
| E6 (V6->V1) | H | C | C | C | C | C | 16.7% |
| E7 (V6->V2) | H | N | H | N | N | N | 33.3% |
| E8 (V6->V4) | H | N | H | H | N | H | 66.7% |
| E9 (V6->V7) | H | N | H | C | C | C | 33.3% |
| E10 (V6->V1) | C | C | N | H | C | H | 33.3% |
| E11 (V7->V1) | H | H | H | H | C | H | 83.3% |
| E12 (V7->V4) | C | C | H | C | H | H | 50% |
| E13 (V8->V1) | H | H | H | H | H | C | 83.3% |
| E14 (V8->V3) | H | N | N | H | N | H | 50% |
| E15 (V8->V7) | H | H | N | H | C | C | 50% |
| E16 (V9->V3) | H | N | N | H | N | C | 33.3% |
| E17 (V9->V7) | H | C | H | H | H | C | 66.7% |
| E18 (V9->V6) | C | N | N | H | H | H | 50% |
| **LLM hallucination rate** | 72.2% | 27.8% | 55.6% | 66.7% | 27.8% | 50% | **Average 50%** |

In total, 108 experiments were conducted for the hallucination survey. We then evaluated the causal debate chart for each experiment to draw a conclusion about the LLMs' performance. After determining a conclusion for each of the 108 experiments, we then calculated the hallucination rate per LLM and per edge.

- For each LLM, we calculated the percentage of edges for which there was a hallucination: ($N_{HE}/N_E \cdot 100\%$), where $N_{HE}$ is the number of such edges and $N_E$ is the total number of edges.
- For each edge, we calculated the percentage of LLMs that hallucinated: ($N_{HLLM}/N_{LLM} \cdot 100\%$), where $N_{HLLM}$ is the number of hallucinating LLMs and $N_{LLM}$ is the total number of LLMs.

Table 1 shows the results of the hallucination survey. We found that out of all 108 experiments conducted, both the LLM hallucination rate and the edge hallucination rate were each 50% on average. This means that regardless of edge, the LLMs hallucinated half the time! Per model, the best performers were GPT-4 and Gemini 1.5 Pro, with a hallucination rate of 27.8% each. The weakest performer was GPT-3.5, with a hallucination rate of 72.2%.

As shown from the wide spectrum of results and performances among the LLMs in the hallucination survey, the LLM used for causal discovery is very important. Since every LLM has different pre-trained knowledge, it makes it more essential for researchers to be able to experiment with different LLMs depending on the problem they are trying to solve. For this particular dataset on life expectancy, for instance, it is possible that GPT-4 and Gemini 1.5

Pro had more pre-trained knowledge on epidemiology, making them have the lowest hallucination rate compared to the other LLMs. By this definition, we construe these LLMs (GPT-4 and Gemini 1.5 Pro) as having the *best performance*.

## 5 Retrieval Augmented Generation

Given the hallucination rates observed across the spectrum of LLMs surveyed, and the strong likelihood of these popular LLMs being employed in causal discovery, researchers should have a way to overcome the hallucination problem observed in these LLMs. We propose using RAG, when quality data exists, for its aforementioned benefits. In the field of causality, no corpus can be more authoritative than the true causal graph if it exists. Therefore, for the life expectancy causal graph in Fig 1., we simulate the true causal graph by creating a textual version where for each edge, we indicate the causality of the edge between the variables as represented by the nodes. For instance, for the edge *primary_care_physicians_rate* → *life_expectancy*, the simulated true causal graph will state: "primary care physicians rate strongly affects life expectancy."

Our objective here is not to construct the true causal graph, but to determine if when given quality data, RAG can be a useful option to reduce LLM hallucinations for auditing causal discovery. We use RAG to augment each LLM with the mocked true causal graph and repeat the 108 experiments from the hallucination survey. We observe the following results in Table 2. Note that we share only the LLM hallucination rate for brevity as we are interested only if RAG improved the hallucination rate of each LLM in this experiment; whether edge hallucination rate improved for any edge or not is irrelevant since the true causal graph was mocked.

Table 2: RAG Survey Results

| Treatment/ LLM | GPT-3.5 | GPT-4 | Llama3 8b | Mixtral 8x7b | Gemini 1.5 Pro | Claude 3.5 Sonnet | **Average LLM hallucination rate** |
|---|---|---|---|---|---|---|---|
| Before RAG | 72.2% | 27.8% | 55.6% | 66.7% | 27.8% | 50% | 50% |
| After RAG | 5.6% | 5.6% | 27.8% | 16.7% | 5.6% | 5.6% | 13.9% |
| **Improvement** | 66.6% | 22.2% | 27.8% | 50% | 22.2% | 44.4% | **Average 36.1%** |

After conducting the RAG survey, the average LLM hallucination rate dropped from 50% in the hallucination survey to 13.9%, demonstrating that RAG was successful in substantially reducing hallucination across the LLMs for causal discovery in general. Individually, GPT-3.5 experienced the most improvement, followed by Mixtral 8x7b and Claude 3.5 Sonnet. Per model, the best performers were GPT-3.5, GPT-4, Gemini 1.5 Pro, and Claude 3.5 Sonnet with a hallucination rate of 5.6% each. The weakest performer was Llama3 8b with a hallucination rate of 27.8%. As an example, Fig. 6–7 shows the success of RAG on the primary_care_physicians_rate → life_expectancy edge on Mixtral 8x7b. As shown by the results, RAG should be an option to researchers when auditing a causal graph with any LLM.

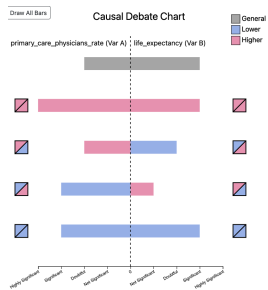
Figure 6: A hallucinated edge before RAG from Mixtral 8x7b.

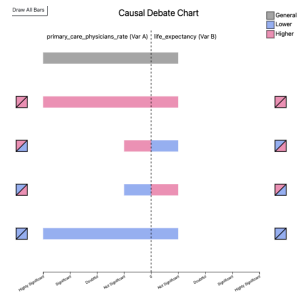
Figure 7: Causality in the same edge after RAG from Mixtral 8x7b.

## 6 Multi-LLM Debate With Arbiter

RAG doesn't eradicate the need for new, quality corpus and these are often elusive, especially in causal discovery, so we simulated quality data by mocking the true causal graph in the RAG experiment. To help overcome this in practice, we propose an alternate approach requiring no new corpus by employing multiple LLMs in a debate. The idea is that we can combine the wealth of knowledge from multiple LLMs to overcome the deficit of any individual LLM and/or lack of quality corpus. Specifically, we create the roles of two debaters, fulfilled by two LLMs, and the role of an arbiter, which is fulfilled by a third LLM. All LLMs are unique from each other. The motion of the debate are the prompts we used to audit our life expectancy causal graph. The debating LLMs will respond to the prompts and we make the arbiter LLM evaluate the responses from both debating LLMs to form the final response to the prompt. To ensure that we create a real debate with opposing views, we made use of the efficiency of RAG and engineered the RAG corpus of the debaters in the following way to ensure that the debater LLMs take up clear Proposition and Opposition positions with strong opinions:

**Proposition**: A strongly affects B, but B does not affect A.
**Opposition**: B strongly affects A, but A does not affect B.

The arbiter required a new prompt, namely:

"Using the responses of the two debaters and your own knowledge, generate a final rating in response to the question: On a scale from 1 to 4, 4 represents strong or most likely, rate the cause-and-effect relationship: {input}"

Note that augmenting the debater LLMs through RAG is unnecessary outside the scope of the experiment as we merely wanted to simulate the worst case scenario when the debaters have conflicting views and observe how the arbiter would synthesize the final response. In the real world where we have a fair debate, every opinion, however opposing or not, is invaluable so that we can synergize the input and wealth of knowledge from each LLM and eliminate the outsized influence of any one LLM or corpus. Hence, we will skip the RAG step when we use the debate method to actually audit a causal graph in practice.

Since the debater LLMs had to be augmented, we selected GPT-4, Gemini and Claude for the debate as they had the best RAG performance from Table 2. Even though GPT-3.5 also had the same RAG performance, we chose to omit GPT-3.5 as it is an older version than GPT-4. We rotated the debaters to compare the permutations. As the arbiters had to moderate the debate to form the final response, they should not be augmented so as to maintain their neutrality. Only Gemini and GPT-4 could fulfill this role as they had the lowest hallucination rate without RAG from Table 1. The following shows the results of the debate experiment.

TABLE 3. Multi-LLM Debate with Arbiter Survey Results

| Debate/Lineup | Proposition | Opposition | Arbiter | LLM hallucination rate |
|---|---|---|---|---|
| 1 | Claude | GPT-4 | Gemini | 11.1% |
| 2 | GPT-4 | Claude | Gemini | 5.6% |
| 3 | Gemini | Claude | GPT-4 | 16.7% |
| 4 | Claude | Gemini | GPT-4 | 16.7% |
| | | | | **Average** 12.5% |

Even though we conducted only one round of debate in each experiment, hallucination rates dropped to levels comparable to those in RAG. For instance, the average in RAG was 13.9% and the debate was actually better at 12.5%. Debate 2's lineup of LLMs and debate roles achieved the same best performance of 5.6% as any individual LLM after RAG. Within each debate, the choice of debaters matters. For instance, in debates 1 and 2, the arbiter was Gemini but when we swapped the roles of Claude and GPT-4, the hallucination rate was different. The choice of arbiters also matters. Debates with Gemini as arbiter had lower hallucination rates than GPT-4 in general but it is more probable that the choice of arbiter and debaters in tandem rather than isolation influenced the hallucination rate of each debate. Our results show that when LLMs are used to audit causal graphs, and limitations with RAG exist (such as lack of quality corpus), the debate approach is an effective alternate option to reduce hallucinations in LLMs. Each debate also had a lower hallucination rate than any LLM initially on its own in Table 1.

## 7 Conclusions

We present the first hallucination survey for LLMs in causal discovery, and show that LLMs are prone to hallucinations, specifically, logical inconsistencies. Since logic is so fundamental to causal discovery, trusting one LLM alone when auditing casual graphs is dangerous given the hallucination frequency of LLMs that we observed but the importance of the problems that causal graphs try to solve. Thus, the choice of LLM in causal discovery must be judicious. Knowing the hallucination rate of a spectrum of commonly used LLMs will be incredibly helpful to researchers in this regard.

We introduce a way to reduce the hallucination rates of LLMs in causal discovery through Retrieval Augmented Generation. However, as RAG depends on quality corpus which can be difficult to find, so it's important to have another viable tool. This is where a debate between LLMs with an arbiter is invaluable, as we show that combining the knowledge of multiple LLMs and the use of an arbiter that adjudicates the debate can reduce hallucinations like RAG.

Our future work may focus on more quantitative ways to define and measure hallucinations, in addition to the causal debate chart, when auditing causal graphs. Also, even though only one round of debate was necessary to observe the reductions in hallucinations, it does not imply that the debate approach was accurate in predicting causality between the variables in the edge. One validation would be to use the debate method on a dataset for which there is a true causal graph and compare the debate response for each edge to evaluate the accuracy of the method. We expect enhancements as those employed by Du et. al. [4], such as more debate rounds, would be necessary to improve our approach. A novel idea may be to create more than one arbiter role like a jury in a court of law and let multiple arbiter LLMs perform a second level of debate in order to construct the final response.

Finally, repeating the experiments with more datasets is paramount to show that the debate method generalizes. Repeating the experiments would also reduce statistical issues such as sampling error and bias. Automation is important to scale this sort of research so using an orchestrator such as LangChain to implement a LLM-assisted causal discovery audit pipeline for researchers can be one useful improvement.

### Acknowledgement

I would like to thank the Simons Foundation for awarding me the opportunity to work on this research at Stony Brook University. I also convey my immense gratitude to Professor Klaus Mueller and Yanming Zhang for their guidance, feedback and encouragement throughout my research.